\documentclass[conference]{IEEEtran}
\IEEEoverridecommandlockouts
\usepackage{cite}
\usepackage{amsmath,amssymb,amsfonts}
\usepackage{algorithmic}
\usepackage{graphicx}
\usepackage{textcomp}
\usepackage{xcolor}
\def\BibTeX{{\rm B\kern-.05em{\sc i\kern-.025em b}\kern-.08em
    T\kern-.1667em\lower.7ex\hbox{E}\kern-.125emX}}
\begin{document}

\title{Deep Bayesian Active Learning, A Brief Survey on Recent Advances \\
}

\author{\IEEEauthorblockN{ Salman Mohamadi}
\IEEEauthorblockA{\textit{Computer Science and Electrical Engineering} \\
\textit{West Virginia University}\\
Morgantown, WV, USA \\
Sm0224@mix.wvu.edu}
\and

\IEEEauthorblockN{Hamidreza Amindavar}
\IEEEauthorblockA{\textit{Electrical Engineering} \\
\textit{Amirkabir University of Technology}\\
Tehran, Iran \\
  hamidami@aut.ac.ir }
}

\maketitle

\begin{abstract}
Active learning frameworks offer efficient data annotation without remarkable accuracy degradation. In other words, active learning starts training the model with a small size of labeled data while exploring the space of unlabeled data in order to select most informative samples to be labeled. Generally speaking, representing the uncertainty is crucial in any active learning framework, however, deep learning methods are not capable of either representing or manipulating model uncertainty. On the other hand, from the real world application perspective, uncertainty representation is getting more and more attention in the machine learning community. Deep Bayesian active learning frameworks and generally any Bayesian active learning settings, provide practical consideration in the model which allows training with small data while representing the model uncertainty for further efficient training. In this paper, we briefly survey recent advances in Bayesian active learning and in particular deep Bayesian active learning frameworks.

\end{abstract}


\section{Introduction}
\label{sec:intro}
In real life application, while data collection may not be as costly and laborious as it was a few decades ago, there are still a lot of considerations that make the data annotation process costly and inefficient for actual deployment of many machine learning algorithms. Therefore, experiment design and particularly, incorporating active learning setting into many of machine problem domains are increasingly gaining attention. Active learning is a framework in the area of machine learning in which the model starts training by small amount of labeled data and then, in a sequential process asks for more data samples from a pool of unlabeled data to label and incorporate in the training process. In fact, the key idea behind this framework is to achieve desired accuracy while lowering the cost of labeling by efficiently asking for more labeling most informative data samples. Therefore, compared to many other relevant frameworks, active learning tries to incorporate the uncertainty representation to achieve the same or higher accuracy by using smaller amount of labeled data. Other than the uncertainty coming from noisy data samples, there are two major types of uncertainty, the uncertainty associated with the best model parameters, and the uncertainty associated with best network structure \cite{b1,b2}. In fact we might find multiple models and/or different structures which provide well representation for the data, however, we may face uncertainty on selecting the one with highest performance on further predictions or generalization. In contrast, similar frameworks such as semi-supervised learning addresses relatively similar problem domains, however, there are differences between their learning paradigm. In more detain, semi-supervised learning frameworks use the  unlabeled data for feature representations in order to better model the labeled data \cite{b3,b4,b5}. Classical methods and tools in signal processing area, with an emphasis on parametric modeling, have been used in many areas with different type of data \cite{b7,b9,b8}. However recent advances in machine learning and in particular, artificial neural network, have shown that non-parametric models, are capable of almost modeling any type of data at the cost of higher complexity. In this line, deep learning could be incorporated with classical tools and frameworks of machine learning such as active learning in order to address a wider range of problems while improving the performance. 

On the other side, representing the uncertainty of either embedding space or output probability space is challenging where we are going to use deep learning tools and concepts. It would show up in multiple scenarios in which we need to measure the model uncertainty such as problems addressed by classical active learning or its various versions with the similar learning paradigm. In fact, exploiting model uncertainty is essential for several problem domains concerning with learning from small amount of data. For instance, it is necessary to develop a desired output probability space in a typical active learning framework, which necessitates the model uncertainty measurement and representation\cite{b28}. Bayesian methods, here can play an important role to capture the underlying model uncertainty. Original idea of incorporating active learning with neural networks has been proposed and assessed few decades ago \cite{b25}, however in recent years, more specific efforts were performed to introduce deep learning tools to active learning framework. In this regard, in order to better understand the addressed problem domain, it is crucial to ponder upon the associated difficulties. Authors of \cite{b6}, in their work as a pioneered effort, discussed that bringing deep learning tools into active learning setting poses two major problem; uncertainty representation and the amount of data possibly needed to train the model. In next two sections, in order to have a taste of the basic concepts and some theoretical definitions, a brief overview of the learning paradigm of active learning and Bayesian convolutional and recurrent neural networks will be presented.

\begin{figure}[t]
	\centering{
		\includegraphics[scale=.425]{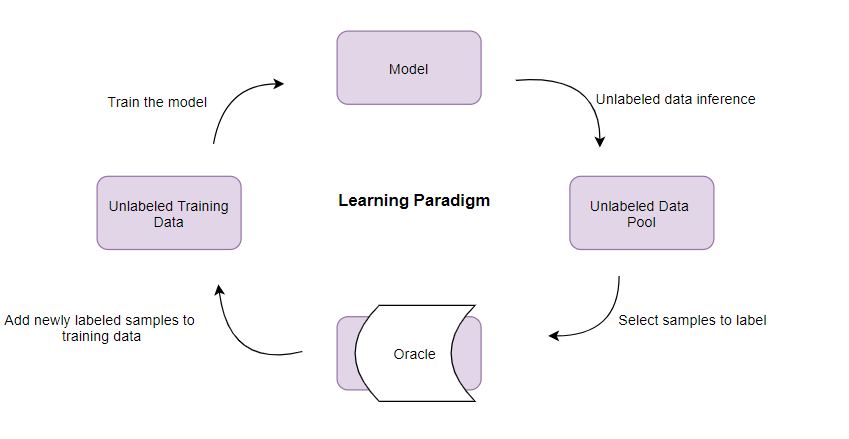}
		\caption{Learning paradigm of active learning; as it is shown, at every iteration the training starts from scratch on modified training data.}
		\label{LP}}
\end{figure}

\section{Learning Paradigm in Active Learning}
Simply put, the goal of active learning is to minimize the cost of data annotation or labeling by efficiently selecting the unlabeled data to be labeled. In more detail, in every iteration of active learning, a new labeled data sample (or even batch of data) will be added to the training data, and the training process starts from scratch. This sequential training will continue until either the accuracy reaches to the desired level, or the entire labeling budget be used.The overview of the learning cycle of active learning is shown in Fig.\ref{LP}.

In each iteration, all of the unlabeled data samples will be evaluated to select the most informative sample. Selection of these samples is performed by a functions named acquisition function which deals with classification uncertainty. The uncertainty presented in classification could be measured in terms of predictive entropy, variation ration and mutual information \cite{b26,b27,b28}. Generally speaking, we make use of acquisition function in active learning framework either for regression or classification purposes. In case of regression, sample variance is the criterion for building and acquisition function. However in case of classification, acquisition functions could be designed based on, maximum predictive entropy-based acquisition,acquisition based on maximum mutual information for predictions and model posterior, maximum variation ratios acquisition or simply random acquisition as a  baseline \cite{b28,b29}. Hence, we could think of acquisition functions as a function performing uncertainty sampling, or diversity sampling or both of them. For the task of image classification, most popular functions are presented and approximately formulated in \cite{b6}.

\section{Projecting Uncertainty in the Models }
\subsection{Convolutional Neural Networks with Bayesian Prior}
Nowadays, deep learning algorithms mostly rely on training convolutional neural networks (CNNs). In fact with the recent advancement of CNNs,  one of the main advantages of CNNs is that they enable capturing spatial information of the image data \cite{b10}. However, Bayesian learning concepts initially were introduced to simple versions of neural networks. In fact first attempt on developing a special type of neural networks (NNs) named Bayesian NNs, dates back to a work presented by reference \cite{b30} more than three decades ago. Their goal was to put a prior distribution over the weights of NN in order to develop a mapping framework between each specific setting of weights and its corresponding sets of outputs. This idea gradually evolved into more complex learning frameworks such as Bayesian CNNs. In essence, practical implementation of Bayesian CNNs requires the measurement of the model predictive posterior on the test data. Theoretically speaking, first each kernel of CNN is set to follow a prior distribution and next, a Bernoulli variational distribution is applied to the resultant kernel-patch pair of the image; in other words the product of Bernoulli random variables and weight matrix is applied to each patch of the image individually. Therefore, some of the patches of image could be multiplied by kernels set to zero. In terms of practical implementation though, it is approximately implemented by performing dropout after every convolutional and fully-connected layer \cite{b28,b11}.

Based on above discussion, authors of \cite{b11} proposed a new version of CNNs with Bayesian prior on a set of weights,i.e., Gaussian prior ${p(w); w }$ is ${ \{W_1, W_2, ...W_N\}}$. In their work, Bayesian CNNs for classification tasks with a softmax layer is formulated with a likelihood model as follows:
\begin{eqnarray}
			\label{equ1}
			p(y=c|x,w)=softmax(f^{w}(x)).
		\end{eqnarray}
As above discussed briefly, in order to practically implement such models, the Gal et.al \cite{b6} suggest approximate inference using stochastic regularization techniques such as dropout or multiplicative Guassian noise. They performed it by utilizing dropout during the training as well as the test process to estimating the posterior over the test process. In more detail, such work is feasible by finding a distribution, namely ${q_{\theta}^*(w)}$ which given a set of training data ${D}$, minimizes the Kullback-Leibler (KL) divergence between estimated posterior and exact posterior ${p(w|D)}$. Finally, as it is common, using Mont Carlo integration for such variational inference, we will have:
\begin{eqnarray}
			\label{equ2}
			p(y=c|x,D) =\int p(y=c|x,w)p(w|D)dw\\
			\approx \int p(y=c|x,w) q_{\theta}^*(w)dw \\
			\approx \frac{1}{T}\sum_{t=1}^{T}p(y=c|x,\hat{w}_t),
		\end{eqnarray}

with ${q_{\theta}(w)}$ as dropout distribution and ${\hat{w}_t}$ as estimation of ${q_{\theta}^*}$ \cite{b6}.

\subsection{Recurrent Neural Networks with Bayesian Prior}
Similar to what is performed on CNN to make it into Bayesian CNN, recurrent neural networks (RNNs) could be modified in terms of model uncertainty.

\section{Review on Recent advances }
Bayesian inference methods allow the introduction of probabilistic framework to machine learning and deep learning. The notion behind the introduction of these kind of frameworks to machine learning is that learning from data would be treated as inferring optimal or near optimal models for data representation, such as automatic model discovery. In this sense, Bayesian methods and here, specifically Bayesian active learning methods gain attention due to their ability for uncertainty representation and even better generalization on small amount of data \cite{b20}.
One of the main work on introduction of model uncertainty measurement and manipulation to active learning is done by Gal and Ghahramani \cite{b6}. In fact the major contribution of this paper is special introduction of Bayesian uncertainty estimation to active learning in order to form a deep active learning framework. In more detail, deep learning tools are data hungry while active learning tends to use small amount of data, moreover, generally speaking deep learning is no suitable for uncertainty representation while active learning relies on model uncertainty measurement or even manipulation. Understanding these big natural differences, authors of this paper found the Bayesian approach to be the solution. In fact they refine the active learning general framework, which usually work with SVM and small amount of data, to be well scaled to high dimensional data such as images in the case of big data. in contrast to small data. It practice, the authors put a Bayesian prior on the kernels of a convolutional neural network as the training engine of active learning framework. They refer to their previous work \cite{b21} suggesting that in order to have a practical Bayesian CNN, the Bayesian inference could be done through approximate inference in the Bayesian CNN, which makes the solution computationally tractable. The interesting point is that they empirically showed that dropout is a Bayesian approximation which can be used as a way to introduce uncertainty to deep learning \cite{b22}. Here the point is that dropout is not only used in the training process, i.e.,  they do inference applying dropout before every weight layer during training, and also during test to sample from the approximate posterior. This framework compared to other active learning methods addressing big data for image, such as those using RBF, performs better.

In this line, Jedoui et. al. \cite{b23} even go further in the level of uncertainty of model by assuming that the output space is no longer mutually exclusive, for instance we have more that one output for a single input. They empirically show that classical uncertainty sampling does not perform better than random sampling at these sort of tasks such as Visual Question Answering, therefore they refer to \cite{b6,b18,b21,b19} take a similar strategy by using Bayesian uncertainty in a semantically structured embedding space rather than modeling uncertainty of the output probability space.  Referring to Gal and Ghahramani's works, they mention that dropout can be interpreted as a variational Bayesian approximation \cite{b21,b22}, where the approximating distribution is a mixture of two Gaussians with small variances which the mean of one of the Gaussians is zero. The prediction uncertainty caused by uncertainty in the weights which could be measured by approximate posterior using Monte Carlo integration. 

Authors of reference \cite{b24} poses another similar problem by introducing deep learning with relatively very large amount of data and big network into active learning; and suggesting the necessity of systematic request for labeling in the form of batch active learning (batch rather than sample in each active learning iteration). They offer batch active learning in order to address the problem that existing greedy algorithms become computationally costly and sensitive to the model slightest changes. The authors propose a model aimed at efficiently scaled active learning by well estimating data posterior. They suggest scenarios in which more efficiency comes with one batch rather than one data sample at each iteration. In this paper, authors take multiple active learning methods, different acquisition functions, into account for their objective of efficient batch selection in the sense of sparsity, or sparse subset approximation. Moreover, they claim that based on their experiments, that reference \cite{b6}, as a Bayesian approach, outperforms others in many problem setting. More specifically, with the same Bayesian active learning framework proposed by \cite{b6} for capturing uncertainty, they target the most optimum batch selection by finding data posterior, however as active learning setting does not provide access to the labels before querying the pool set, they take expectation w.r.t. the current predictive posterior distribution. 
This work represents a closed-form solution consistent with basic theoretical setting of reference \cite{b6}.

Gal and Ghahramani \cite{b11} suggest a Bernoulli approximate variational inference method, which prevents from CNNs over-fitting, i.e., by considering a Bayesian prior on the weights of the network, it will become capable of learning from small data, with no over-fitting or higher computational complexity. This work can be considered as one of the basics of developing deep Bayesian active learning. 
Authors of \cite{b12} with an emphasis on the fact that the nature of active learning does not allows thorough comparison of models and acquisition functions, explore more than 4 experiments of different models and acquisition functions for multiple tasks of natural language processing, and finally show that deep Bayesian active learning consistently provides the best performance. Kandasamy et. al\cite{b13} underscore the fact that classical methods for posterior estimation are query inefficient in the sense of estimating likelihood. They suggest that a query efficient approach would be posterior approximation using Bayesian active learning framework. Considering a Gaussian prior, a utility function as a measure of divergence between probabilities (here densities)is formed and at each time step, the estimated most informative query would be sent to an oracle. Then the posterior will be updated. Their experiment confirms the query efficiency of the approach. Reference \cite{b14} suggests that since most of the methods of distance metric learning are sensitive to the size of the data and randomly select the training pairs, they could not return satisfactory results in many real world problems. The authors address this problem by firstly introducing Bayesian approach to distance metric learning, and then developing this framework by uncertainty modeling, which ends up in a Bayesian framework for active distance metric learning. This framework enables efficient pair selection for training as well as   posterior probability approximation. Reference \cite{b15} tries to combine the benefits of Bayesian active learning and semi-supervised learning by introducing a active expectation maximization framework with pseudo-labels. Authors use Mont Carlo dropout introduce in \cite{b6} to compute the probability outputs. 

\cite{b32,b31} present innovative strategies for addressing the issues of sampling and visualizing high-dimensional unbalanced datasets to reduce computing complexity and memory utilization while preserving accuracy. This could be used as an auxiliary technique for deep active learning frameworks. 

Zeng et. al \cite{b16} address the question that is it possible to measure the model uncertainty without fully Bayesian CNNs. Their results on several Bayesian CNNs confirm that in order to represent the model uncertainty, one needs to apply the Bayesian prior on only a few last layer before the output. With this setting, the model would enjoy the benefits of both deterministic and Bayesian CNNs. Lewenberg et. al \cite{b17} address the problem of active surveying using a Bayesian active learner. In fact they use dimensionality reduction in an active learning framework by applying Bayesian prior in order to design a system to predict the answers to unasked question using a limited sequential active question asking. Their framework outperform several state of-the-art frameworks based on enhanced linear regression in terms of prediction accuracy and response to missing data.

Houlsby et. al.\cite{b18} propose a measurement of predictive entropy which later is used in a classification framework based on Gaussian Process. Their method performs well compared to several similar classification frameworks while the computational complexity is not greater than other methods. Finally they develop their framework to a Gaussian process preference learning  by extension of binary preference learning to classification setting. One of the main advantages of this method is that it provides desired accuracy at a relatively low computational complexity cost. 

\section{Future trends}

\section{Conclusion}
	
	In this paper, we surveyed recent advances in Bayesian active learning, with an emphasis on deep Bayesian active learning. Our main focus in on the works contributing to the theory of this problem domain, however some interesting works on the application of Bayesian active learning are surveyed. Bayesian inference approaches hold very important place in machine learning and recently, many attentions have shifted toward data, model and even network structure uncertainty representation using these approaches. Bayesian active learning and its intersection with deep learning concepts provide very interesting frameworks in terms of theroy and application.


\end{document}